\documentclass[11pt]{article}

\usepackage[T1]{fontenc}
\usepackage[letterpaper,margin=1in]{geometry}
\usepackage{microtype}
\usepackage{amsmath,amssymb}
\usepackage{graphicx}
\usepackage{booktabs}
\usepackage{enumitem}
\usepackage{float}
\usepackage[ruled,vlined]{algorithm2e}
\usepackage{xspace}
\usepackage{multirow}
\usepackage{multicol}
\usepackage{subcaption}
\usepackage[table]{xcolor}
\usepackage{cite}
\usepackage[hidelinks]{hyperref}
\usepackage{orcidlink}
\usepackage[capitalize,nameinlink]{cleveref}

\hypersetup{
  pdftitle={Detecting Backdoors in Object Detection via Pre-NMS Prediction Distribution Shift},
  pdfauthor={Longtian Wang, Zhengyu Zhao, Chenhao Lin, Le Yang, Shiwei Wang, Yuhan Zhi, Xiaofei Xie, and Chao Shen}
}
\makeatletter
\DeclareRobustCommand\onedot{\futurelet\@let@token\@onedot}
\def\@onedot{\ifx\@let@token.\else.\null\fi\xspace}

\def\ie{\emph{i.e}\onedot}

\makeatother

\newcommand{\keywords}[1]{\par\medskip\noindent\textbf{Keywords:} #1}
\newcommand{\tool}{\textit{DistScan}\xspace}

\begin{document}

\title{Detecting Backdoors in Object Detection via Pre-NMS Prediction Distribution Shift} 
\author{%
Longtian Wang\textsuperscript{1}\orcidlink{0009-0005-9145-743X},
Zhengyu Zhao\textsuperscript{1}\thanks{Corresponding author: \href{mailto:zhengyu.zhao@xjtu.edu.cn}{\texttt{zhengyu.zhao@xjtu.edu.cn}}},
Chenhao Lin\textsuperscript{1}, Le Yang\textsuperscript{1},\\
Shiwei Wang\textsuperscript{1}, Yuhan Zhi\textsuperscript{1},
Xiaofei Xie\textsuperscript{2}, Chao Shen\textsuperscript{1}\\[0.6em]
\small \textsuperscript{1}Xi'an Jiaotong University, Xi'an, China\\
\small \textsuperscript{2}Singapore Management University, Singapore
}
\date{}

\maketitle

\begin{abstract}
Object detection models deployed in safety-critical applications remain vulnerable to backdoor attacks that cause targeted misbehaviors when a hidden trigger is present. Existing detection methods either rely on trigger inversion or exploit architecture-specific assumptions, and critically, representative existing methods fail to generalize reliably to scene-level attacks, where a single trigger induces anomalous behavior across all objects in the scene simultaneously. We present \tool, a backdoor detection framework based on a simple but previously unexploited observation: backdoor injection systematically shifts a model's pre-NMS prediction class distribution away from its training class frequencies, even on clean inputs without any trigger present. \tool aggregates intermediate class predictions over a clean validation set and flags a model as backdoored if the resulting distribution deviates significantly from the training class frequencies, requiring no model weight access, no trigger knowledge, and no additional training. Extensive experiments on MS-COCO and PASCAL VOC across two architectures and three scene-level attack scenarios demonstrate that \tool substantially outperforms existing methods, improving average detection accuracy over the best-performing applicable baseline by 27.32 percentage points.

\keywords{Backdoor scanning, object detection, distribution shift}
\end{abstract}

\section{Introduction}
\label{sec:intro}

Object detection models have been widely deployed in safety-critical applications such as autonomous driving~\cite{jia2023fast, cao2023review, deng2025enhanced}, intelligent surveillance~\cite{alamri2025comprehensive, ouardirhi2024enhancing, abba2024real}, and medical image analysis~\cite{dadjouy2024gallbladder, albuquerque2025deep, saraei2025deep}. In these scenarios, detection results directly influence downstream decision-making and system safety~\cite{ceccarelli2023evaluating}. However, recent studies have shown that object detection models are vulnerable to backdoor attacks~\cite{chan2022baddet, detectorcollapse2025}. By poisoning the training process, an adversary can implant a hidden backdoor such that the model behaves maliciously when a specific trigger is present, for example, failing to detect pedestrians or misidentifying prohibited items in security screening, while maintaining near-normal detection performance on benign inputs. This stealthy behavior makes backdoored models difficult to detect through conventional validation procedures before deployment, posing serious security risks in real-world applications. Therefore, effective and practical backdoor detection for object detection models is of critical importance.

\begin{figure}[H]
    \centering
    \begin{subfigure}{0.46\linewidth}
        \includegraphics[width=\linewidth]{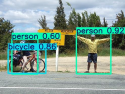}
        \caption{Without Trigger}
    \end{subfigure}
    \hfill
    \begin{subfigure}{0.46\linewidth}
        \includegraphics[width=\linewidth]{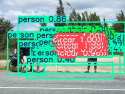}
        \caption{With Trigger}
    \end{subfigure}
    \caption{A Scene-Level Backdoor Attack Triggering Numerous Spurious Detections.}
    \label{fig:scene_level_attack}
\end{figure}

Although backdoor detection has been extensively studied in image classification~\cite{gu2019badnets,chen2017targeted, li2021invisible, nguyen2021wanet, zeng2021rethinking}, these methods have been shown to suffer significant performance degradation when transferred to object detection, as the two tasks differ fundamentally in both model structure and learning objective~\cite{odscan2024}. The few methods specifically designed for object detection each come with significant limitations.
Specifically, trigger reconstruction-based approaches such as ODSCAN~\cite{odscan2024} require prior assumptions about trigger form to constrain the search space, as the optimization becomes intractable when trigger size, shape, and style are unknown. Inconsistency-based approaches such as MIA~\cite{mia2024} exploit architecture-specific signals that are inherently restricted to two-stage detection. More critically, both approaches fail against scene-level attacks such as Detector Collapse (DC)~\cite{detectorcollapse2025}, where a trigger induces simultaneous detection anomalies across the entire scene as shown in~\cref{fig:scene_level_attack}. ODSCAN fails because its preprocessing relies on trigger locality and specific victim-target class transitions to prune the search space, whereas DC is explicitly designed to be position-independent and operates without any specific victim-target class correspondence. 
MIA is limited because DC simultaneously corrupts both the regression and classification branches~\cite{detectorcollapse2025}, which eliminates the inter-module divergence that the method depends on for detection.

To address these limitations, we seek a detection approach that requires no prior assumption of the trigger, generalizes across detector architectures, and is effective against scene-level attacks. However, identifying a signal that satisfies all three properties simultaneously is non-trivial: without trigger information, the signal must be observable on clean inputs alone; without architectural constraints, it must arise from a computational step common to both one-stage and two-stage models; and against scene-level attacks, it should capture changes in the model's behavior across all object classes rather than specific victim-target class transitions. Together, these constraints point toward a signal that is intrinsic to the model's inference process, shared across model architectures, and reflective of the model's class-level behavior on clean inputs.

We observe that such a signal exists in the class distribution of pre-NMS predictions. A benign model's pre-NMS predictions closely align with the class distribution of the training data, as the prediction-generating components are optimized directly on the training data and implicitly encode its class-frequency prior~\cite{torralba2011unbiased, papyan2020prevalence}. Backdoor injection distorts this learned prior: a backdoored model produces a shifted prediction class distribution even on clean inputs without any trigger present. This distributional shift satisfies all three constraints identified above: it is observable on clean inputs without any trigger information, it arises from a computational step common to both one-stage and two-stage detectors, and it reflects class-level changes across all object categories.

Motivated by this observation, we propose \tool, a backdoor detection framework that exploits the pre-NMS prediction class distribution shift as a discriminative signal. To account for the architectural differences between one-stage and two-stage detection models, \tool constructs the validation set based on the architectural characteristics and training paradigms of these two types. \tool then extracts the pre-NMS prediction class distribution produced by the model under inspection and quantifies its divergence from the training class distribution using Jensen-Shannon (JS) divergence~\cite{lin1991divergence}. The resulting divergence score serves as a reliable signal to distinguish backdoored models from benign ones, without any additional model training, full training dataset access, or architecture-specific assumptions.

In summary, this work makes the following contributions:
\begin{itemize}[label=\textbullet, leftmargin=*]
\item \textbf{A novel detection perspective.} To the best of our knowledge, we are the first to identify and characterize the class distribution shift in pre-NMS predictions between backdoored and benign object detection models, and to demonstrate its effectiveness as a reliable signal for backdoor detection.
\item \textbf{An effective detection framework.} We propose \tool, a distribution-based backdoor detection method that requires only a small set of clean samples and the training class distribution, achieving accurate detection without any additional model training, full dataset access, or architecture-specific assumptions.
\item \textbf{Comprehensive experimental validation.} Extensive experiments on two mainstream object detection architectures (YOLOv5 and Faster R-CNN) and two widely used benchmarks (PASCAL VOC and MS-COCO) across three scene-level attack scenarios, comprising 288 object detection models in total, demonstrate that \tool achieves an average detection accuracy of 96.99\%, outperforming the best-performing applicable baseline by 27.32 percentage points on average.
\end{itemize}

\section{Background}
\label{sec:background}

\subsection{Object Detection}
Object detection requires a model to simultaneously localize and classify all objects present in the input. Formally, given an input image $x$, an object detection model $f_\theta(\cdot)$ produces a set of $K$ predictions $\hat{y} = \{(\hat{c}_j, \hat{\mathbf{b}}_j, \hat{s}_j)\}_{j=1}^{K}$, where $K$ is the number of detections, and $\hat{c}_j$, $\hat{\mathbf{b}}_j$, and $\hat{s}_j$ denote the predicted class, bounding box, and confidence score of the $j$-th object detected, respectively.

Modern object detection models fall into two paradigms. Two-stage detection models such as Faster R-CNN~\cite{ren2015faster} first generate class-agnostic region proposals via a Region Proposal Network (RPN), then classify and refine each proposal through a classification head, producing per-class scored predictions that are subsequently filtered by non-maximum suppression (NMS). 
One-stage detection models such as YOLO~\cite{redmon2016you} directly regress bounding boxes and class scores from dense feature maps, similarly producing a large set of candidate predictions before NMS filtering. In both paradigms, we refer to these pre-NMS intermediate predictions uniformly as \textit{pre-NMS predictions} throughout this paper.

\subsection{Backdoor Attacks}

\noindent\textbf{In image classification.}
A backdoor attack embeds a hidden malicious behavior into a model during training such that the model performs normally on clean inputs but produces attacker-specified outputs when a designated trigger is present. The predominant attack vector is data poisoning~\cite{gu2019badnets,chen2017targeted}, and subsequent work has pursued increasingly stealthy triggers including imperceptible perturbations~\cite{li2021invisible}, geometric warping~\cite{nguyen2021wanet}, and frequency-domain modifications~\cite{zeng2021rethinking}.

\noindent\textbf{In object detection.}
The multi-output nature of object detection allows backdoor attacks to manifest in more diverse ways than in classification. BadDet~\cite{chan2022baddet} systematically explores this by proposing attacks targeting object misclassification, object disappearance, and object insertion. More recent scene-level attacks, such as Detector Collapse~\cite{detectorcollapse2025}, demonstrate that a single backdoor can simultaneously induce all three effects across all objects in the entire scene.

\subsection{Backdoor Defenses}

\noindent\textbf{In image classification.}
Existing defenses include trigger inversion methods such as Neural Cleanse~\cite{wang2019neural} and ABS~\cite{liu2019abs}, meta-classifier approaches such as MNTD~\cite{xu2021detecting}, removal methods such as Fine-Pruning~\cite{liu2018fine} and NAD~\cite{li2021nad}, and poisoned-sample detection methods such as PSBD~\cite{li2024psbd}. These methods share an implicit single-label assumption that does not transfer to object detection~\cite{odscan2024}.

\noindent\textbf{In object detection.}
ODSCAN~\cite{odscan2024} adapts trigger inversion to object detection by exploiting structural properties to reduce the search space and handling trigger specificity via polygon region inversion. MIA~\cite{mia2024} detects backdoors via behavioral inconsistency between the RPN and classification head in two-stage models. Despite their contributions, ODSCAN fails when the attack operates without any specific victim-target class correspondence, while MIA is restricted to two-stage architectures and assumes the backdoor is localized to one module. As demonstrated by our experiments, none of these methods generalizes effectively to scene-level attacks.

\section{Methodology}
\label{sec:method}

\subsection{Threat Model}
\noindent\textbf{Attacker's Goal and Capability.} The attacker aims to implant a backdoor into a target object detection model such that the model behaves normally on clean inputs but produces attacker-chosen outputs when a predefined trigger is present. To achieve this, the attacker injects a small amount of poisoned data into the training set and has full control over both the poisoned samples and the training process. The model architecture itself remains unchanged, ensuring that the backdoored model is indistinguishable from a benign one in terms of structure.

\noindent\textbf{Defender's Goal and Capability.}
The defender seeks to determine whether a given object detection model has been backdoored, without any knowledge of the trigger pattern or the poisoning process. This reflects a realistic deployment scenario in which a defender receives a trained model from an untrusted third party and must perform inspection prior to deployment~\cite{trace2025,debackdoor2025}.
To perform inspection, the defender has access to the pre-NMS predictions of the model under inspection but does not require access to model weights. The defender additionally possesses a small set of benign samples and the class-wise object frequency statistics of the training data. The latter is a practically grounded assumption: standardized inspection frameworks such as IARPA TrojAI~\cite{trojaidata} provide per-model dataset statistics alongside benign example images, and many publicly released models document training set composition without releasing raw data~\cite{clip2021, align2021, groundingdino2023}.

\subsection{Key Intuition}
\label{sec:key_intuition}

\begin{figure}[!ht]
\centering
\includegraphics[width=0.72\linewidth]{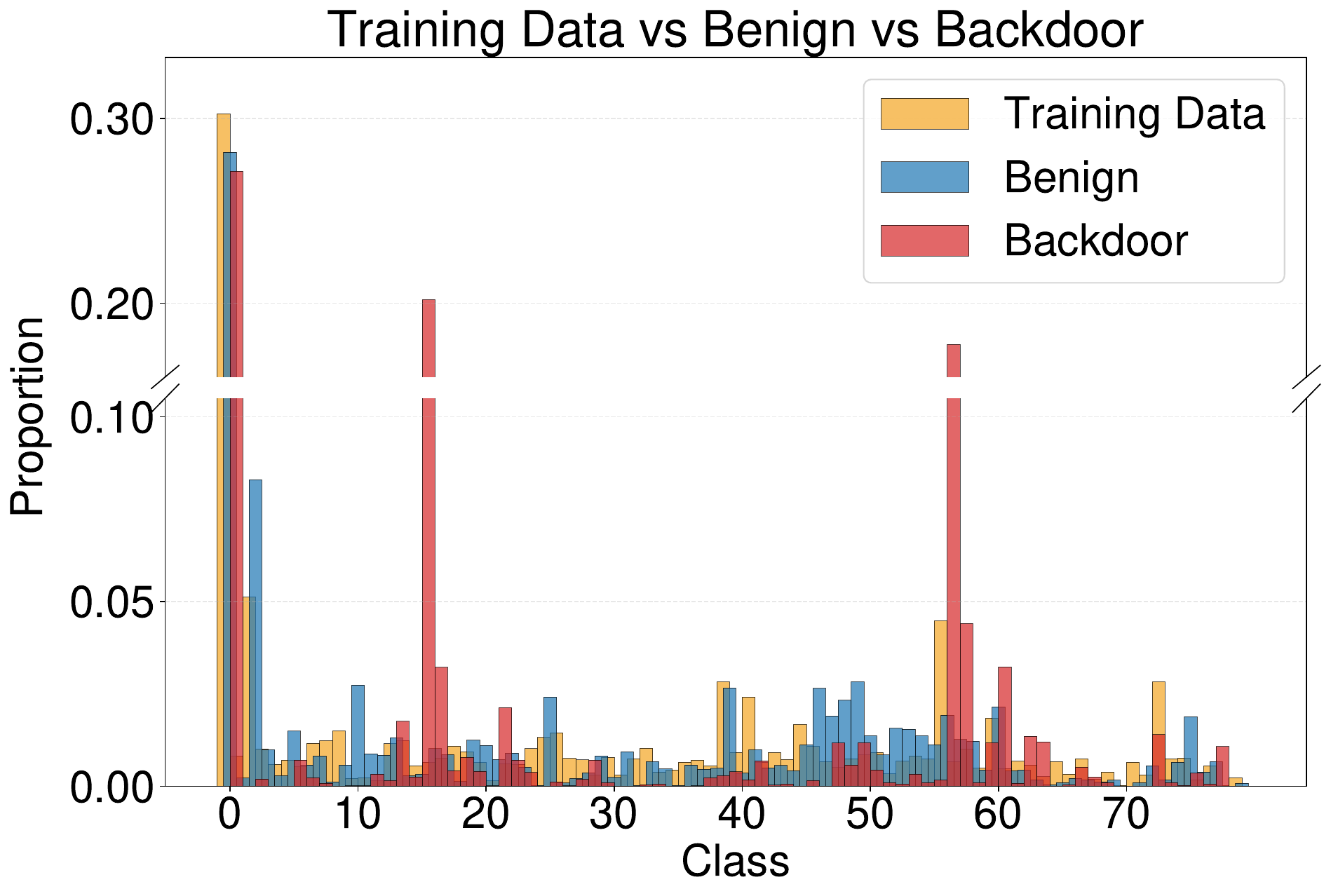}
\caption{Pre-NMS prediction class distributions of benign (blue) and backdoored (red) models, with the training data distribution marked as yellow.}
\label{fig:training_data_distribution}
\end{figure}

Modern object detection models, regardless of paradigm, generate a large number of pre-NMS predictions as an intermediate step before the final output. Since the prediction-generating components are optimized directly on the training data, they implicitly encode the class-frequency prior of the training set, consistent with the well-established observation that models internalize the statistical structure of their training data~\cite{torralba2011unbiased, papyan2020prevalence}. Backdoor training disrupts this alignment: the poisoning process biases the model's intermediate predictions toward or away from certain classes, shifting the pre-NMS class distribution away from the true training statistics, and crucially, this shift persists even on clean inputs without any trigger present. As a result, benign models are expected to produce pre-NMS prediction class distributions closer to the training data than backdoored models, as we empirically verify in~\cref{fig:training_data_distribution} using Faster R-CNN models trained on MS-COCO under a scene-level attack that induces widespread object insertion, where each model's class distribution is estimated by aggregating pre-NMS predictions over 800 randomly selected clean images. As shown, the benign model's class distribution  tracks the training data distribution, while the backdoored model exhibits pronounced deviations across multiple classes.

Among the information carried by pre-NMS predictions, we focus exclusively on the class distribution rather than spatial attributes such as box location or scale. The main reason is that class distributions can be directly grounded against a reference derived from training data statistics, providing a principled and model-independent baseline for comparison, whereas spatial attributes lack such a natural prior, making it difficult to establish a stable reference for detection. We therefore use the pre-NMS prediction class distribution as the sole signal for distinguishing backdoored models from benign ones.

\subsection{Detailed Method}
\begin{figure}[t]
\centering
\includegraphics[width=\textwidth]{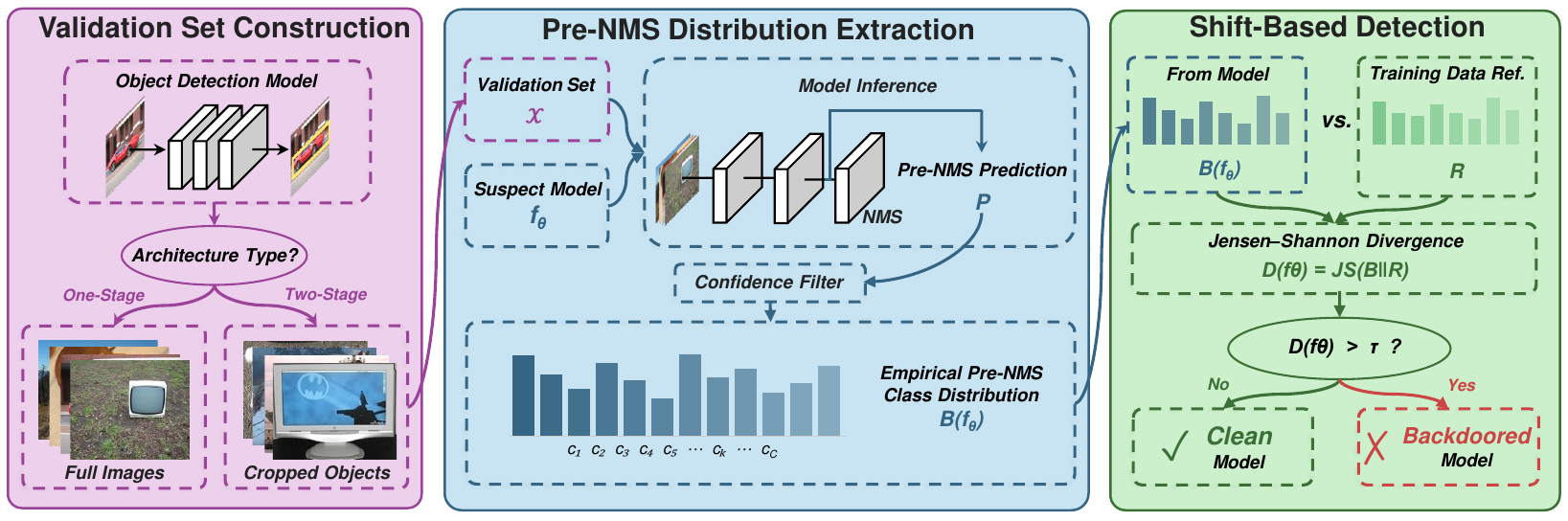}
\caption{Overview of \tool.}
\label{fig:overview}
\end{figure}

\cref{fig:overview} presents an overview of the proposed detection framework. Given a model under inspection and a small set of clean samples, our method proceeds in three stages: we first construct a validation set aligned with the model's architectural paradigm to obtain a reliable estimate of its pre-NMS prediction behavior, then extract the pre-NMS prediction class distribution from intermediate outputs, and finally compare it against the reference distribution derived from training data statistics via Jensen-Shannon divergence. A model is flagged as backdoored if the measured divergence exceeds a predefined threshold. The complete procedure is summarized in~\cref{alg:detection}.

\noindent\textbf{Validation Set Construction.}
As established in~\cref{sec:key_intuition}, the pre-NMS prediction class distribution serves as our detection signal. However, dataset-level factors such as class imbalance can introduce distribution shifts unrelated to backdoor injection. To reduce such confounding effects, we construct the validation set $\mathcal{X}$ to match the input distribution exposed to the detector's classification head, while controlling class composition when beneficial.

The construction of $\mathcal{X}$ must account for the architectural paradigm of the model under inspection, as different detection model designs expose their classification heads to fundamentally different input distributions during training. For one-stage models such as YOLO, the classification head is trained on full images via dense anchor predictions, so we construct $\mathcal{X}$ from complete clean images to match this training distribution. For two-stage models such as Faster R-CNN, the second-stage classification head is trained on RoI-pooled proposal features rather than full scenes, so we instead construct $\mathcal{X}$ using cropped single-object images as a practical proxy that emphasizes foreground-object responses while controlling object scale and class composition. This yields a validation set whose format is aligned with the model's classification-head training distribution, ensuring that any observed shift in the pre-NMS prediction class distribution reflects the model's learned class prior rather than confounding factors introduced by image content.

\begin{algorithm}[t]
\caption{Distribution-based Backdoor Detection}
\label{alg:detection}
\KwIn{Model under inspection $f_\theta$, validation set $\mathcal{X}$, training class counts $\mathbf{R}$, confidence threshold $\delta$, detection threshold $\tau$}
\KwOut{Detection result $\mathrm{Backdoor}(f_\theta) \in \{0, 1\}$}
\nl $\hat{\mathbf{R}} \leftarrow \mathbf{R} / \|\mathbf{R}\|_1$ \tcp{normalize reference distribution}
\nl $\mathbf{S} \leftarrow \mathbf{0} \in \mathbb{R}^C$ \tcp{initialize class count accumulator}
\nl \For{each image $x_i \in \mathcal{X}$}{
\nl     $\mathcal{P}_i \leftarrow$ pre-NMS predictions of $f_\theta$ on $x_i$\;
\nl     $\mathcal{P}_i^{\delta} \leftarrow \{(\hat{c}_j, \hat{\mathbf{b}}_j, \hat{s}_j) \in \mathcal{P}_i \mid \hat{s}_j \geq \delta\}$ \tcp{filter by confidence threshold}
\nl     \For{each prediction $(\hat{c}_j, \hat{\mathbf{b}}_j, \hat{s}_j) \in \mathcal{P}_i^{\delta}$}{
\nl         $\mathbf{S}[\hat{c}_j] \leftarrow \mathbf{S}[\hat{c}_j] + 1$\;
        }
    }
\nl $\hat{\mathbf{B}}(f_\theta) \leftarrow \mathbf{S} / \|\mathbf{S}\|_1$ \tcp{normalize to obtain pre-NMS prediction class distribution}
\nl $D(f_\theta) \leftarrow \mathrm{JS}(\hat{\mathbf{B}}(f_\theta) \| \hat{\mathbf{R}})$ \tcp{compute JS divergence}
\nl \eIf{$D(f_\theta) > \tau$}{
\nl     \Return $1$ \tcp{backdoored}
}{
\nl     \Return $0$ \tcp{benign}
}
\end{algorithm}

\noindent\textbf{Pre-NMS Prediction Distribution Extraction.}
For each image $x_i \in \mathcal{X}$, we feed it through the model under inspection $f_\theta(\cdot)$ and collect the pre-NMS predictions. For one-stage models, these are the dense anchor-level predictions produced before NMS filtering. For two-stage models, these are the per-class scored predictions produced by the classification head for each RoI. In both cases, each prediction is represented as a tuple $(\hat{c}_j, \hat{\mathbf{b}}_j, \hat{s}_j)$, and we denote the full set for image $x_i$ as $\mathcal{P}_i = \{(\hat{c}_j, \hat{\mathbf{b}}_j, \hat{s}_j)\}_{j=1}^{K_i}$, where $K_i$ is the total number of pre-NMS predictions (\cref{alg:detection}, lines~4--8).

To retain only meaningful predictions and eliminate the effect of uniformly distributed low-confidence outputs, we apply a confidence threshold $\delta$ and discard predictions with $\hat{s}_j < \delta$ (line~5). From the remaining predictions, we construct a class-wise count vector
\begin{equation}
    \mathbf{b}_i(f_\theta) = [b_i^1(f_\theta), b_i^2(f_\theta), \dots, b_i^C(f_\theta)],
\end{equation}
where $b_i^c(f_\theta)$ denotes the number of retained predictions for class $c$ produced by $f_\theta$ on image $x_i$. We aggregate these vectors into the accumulator $\mathbf{S}(f_\theta)=\sum_{i=1}^{|\mathcal{X}|}\mathbf{b}_i(f_\theta)$, corresponding to $\mathbf{S}$ in Algorithm~\ref{alg:detection}, and normalize it to obtain the pre-NMS prediction class distribution of $f_\theta$:
\begin{equation}
    \hat{\mathbf{B}}(f_\theta) =
    \frac{\mathbf{S}(f_\theta)}{\|\mathbf{S}(f_\theta)\|_1}.
\end{equation}
We take the class-wise instance counts of the training data $\mathbf{R} = [r^1, r^2, \dots, r^C]$ as the reference, where $r^c$ is the number of annotated instances of class $c$, and normalize to obtain the reference distribution (line~1):
\begin{equation}
    \hat{\mathbf{R}} = \frac{\mathbf{R}}{\|\mathbf{R}\|_1}.
\end{equation}
This normalized distribution serves as the reference against which we compare each model's pre-NMS prediction class distribution, grounded in the alignment between a model's learned class prior and its training data statistics established in~\cref{sec:key_intuition}. Normalization is necessary because the raw training data counts and the prediction counts accumulated over the validation set differ in absolute scale, which would cause distance measurements to be dominated by scale rather than class-level deviation. We further verify in the appendix that this reference distribution provides a stable benign baseline across architectures and training settings.

\noindent\textbf{Shift-based Detection.}
We measure the shift between the pre-NMS prediction class distribution of $f_\theta$ and the reference distribution using Jensen-Shannon divergence (line~9):
\begin{equation}
    D(f_\theta) = \mathrm{JS}(\hat{\mathbf{B}}(f_\theta) \| \hat{\mathbf{R}}).
\end{equation}
The model $f_\theta$ is flagged as backdoored if the divergence exceeds a threshold $\tau$ (lines~10--12):
\begin{equation}
    \mathrm{Backdoor}(f_\theta) =
    \begin{cases}
        1, & \text{if } D(f_\theta) > \tau, \\
        0, & \text{otherwise}.
    \end{cases}
\end{equation}
We use JS divergence because it remains finite when some classes have near-zero probability and provides a symmetric measure for comparing class distributions. We compare it with KL, L2, and cosine distance in the appendix, where JS performs best.

\section{Evaluation}
\label{sec:exp}

\subsection{Experimental Settings}

\subsubsection{Datasets and Models.}
We conduct our evaluation on two widely used object detection benchmarks: PASCAL VOC (VOC)~\cite{everingham2010pascal} and MS-COCO (COCO)~\cite{lin2014microsoft}. For detection architectures, we evaluate on YOLOv5~\cite{jocher_yolov5_2020} as a representative one-stage model and Faster R-CNN~\cite{ren2015faster} with a ResNet-50 backbone as a representative two-stage model, covering both major detection paradigms.

\subsubsection{Evaluation Metrics.}
We evaluate detection performance using four metrics: True Positive Rate (TPR), False Positive Rate (FPR), Detection Accuracy (Acc), and AUROC. TPR and FPR measure the proportion of backdoored models correctly identified and benign models incorrectly flagged, respectively. Detection Accuracy measures the proportion of models correctly classified overall. For each method, the detection threshold $\tau$ is set to the value that maximizes its accuracy on the evaluation model pool, ensuring each method operates at its best possible operating point. AUROC evaluates discrimination performance across all possible thresholds, with higher values indicating more robust separation between backdoored and benign models regardless of the choice of $\tau$. We additionally validate a label-free threshold calibration protocol in the appendix, where $\tau$ is selected from benign models only.

\subsubsection{Evaluation Model Pool Construction.}
To evaluate our method across diverse attack scenarios and model configurations, we construct a pool of backdoored and benign models following a similar construction strategy to MNTD~\cite{xu2021detecting}. For each dataset and architecture combination, we train backdoored models per attack type by randomly sampling combinations of trigger pattern, trigger size, trigger placement, and poisoning rate from a predefined configuration space, ensuring sufficient diversity in the model pool to assess the generalization of our detection method across a wide range of model behaviors. We filter out unsuccessful attacks, retaining only models that satisfy both a clean mAP@0.5 above 0.6 on benign inputs and a backdoored mAP@0.5 below 0.05 on triggered inputs. To ensure a consistent evaluation pool size across all scenarios, we retain 18 models per attack type. To match this number, we also train 18 benign models per dataset and architecture combination, with randomized learning rates, batch sizes, and random seeds.

We consider three attack types, each representing a distinct threat scenario in object detection:
\begin{itemize}[label=\textbullet, leftmargin=*]
    \item \textit{Object Misclassification.} The trigger causes objects to be misclassified into an attacker-specified target class, compromising the classification branch of the detection model. We implement this attack following GMA~\cite{chan2022baddet}, with the target class randomly assigned for each backdoored model.
    \item \textit{Object Disappearance.} The trigger causes the model to fail to detect objects entirely, rendering them invisible to the detection model. We implement this attack following BLINDING~\cite{detectorcollapse2025}.
    \item \textit{Object Insertion.} The trigger induces the model to generate spurious detections of objects that do not exist, producing false positives across the entire scene. We implement this attack following SPONGE~\cite{detectorcollapse2025}.
\end{itemize}

In total, across 3 attack types and 1 benign setting, 2 datasets, and 2 architectures, we obtain $4 \times 2 \times 2 \times 18 = 288$ models for evaluation.

\subsubsection{Implementation Details.}
We construct validation sets as described in~\cref{sec:method}. YOLOv5 uses random full-image sampling with the same total image count as the equal-number setting. Faster R-CNN uses equal-size cropped single-object images with $N=10$ instances per class, resulting in 200 crops for PASCAL VOC (20 classes) and 800 crops for COCO (80 classes). For pre-NMS prediction extraction, we set the confidence threshold to $\delta = 0.0005$ to filter out low-confidence outputs while retaining a sufficient number of meaningful predictions. The effects of the validation set construction strategy and $\delta$ on detection performance are empirically analyzed in~\cref{sec:ablation_valset}.

\subsubsection{Baselines.}
We compare \tool against two representative backdoor detection methods for object detection models. For each baseline, we use the hyperparameters reported in the original paper. ODSCAN~\cite{odscan2024} is a trigger reconstruction approach that reduces the search space via trigger locality and victim-target transition, using polygon region inversion and confidence-aided separation to reconstruct backdoor triggers. MIA~\cite{mia2024} exploits the behavioral discrepancy between the Region Proposal Network and the classification head, flagging a model as backdoored when the mean per-proposal inconsistency score exceeds a predefined threshold.

\subsection{Results}

\subsubsection{Effectiveness of \tool in Detecting Backdoor Models.}

\begin{table}[htbp]
  \centering
  \caption{Comparison of \tool and baselines on detection accuracy, TPR, and FPR. FRCNN denotes Faster R-CNN and Acc. denotes detection accuracy. `-' indicates that the method is not applicable to the given setting. Best results across methods are \textbf{bolded}. }
  \resizebox{\linewidth}{!}{%
    \begin{tabular}{ccc|rrr|rrr|ccc}
    \toprule
    \multirow{2}[4]{*}{Scenario} & \multirow{2}[4]{*}{Dataset} & \multirow{2}[4]{*}{Model} & \multicolumn{3}{c|}{\tool} & \multicolumn{3}{c|}{MIA} & \multicolumn{3}{c}{ODSCAN} \\
\cmidrule(lr){4-6}\cmidrule(lr){7-9}\cmidrule(lr){10-12}
          &       &       & \multicolumn{1}{c}{TPR} & \multicolumn{1}{c}{FPR} & \multicolumn{1}{c|}{Acc.} & \multicolumn{1}{c}{TPR} & \multicolumn{1}{c}{FPR} & \multicolumn{1}{c|}{Acc.} & TPR   & FPR   & Acc. \\
    \midrule
    \multirow{4}[2]{*}{SPONGE} & \multirow{2}[1]{*}{VOC} & YOLOv5 & 100.00 & 0.00 & \textbf{100.00} & - & - & - & 0.00 & 0.00 & 50.00 \\
          &       & FRCNN & 94.44 & 0.00 & \textbf{97.22} & 0.00 & 0.00 & 50.00 & 0.00 & 0.00 & 50.00 \\
          & \multirow{2}[1]{*}{COCO} & YOLOv5 & 100.00 & 0.00 & \textbf{100.00} & - & - & - & 0.00 & 0.00 & 50.00 \\
          &       & FRCNN & 83.33 & 0.00 & \textbf{91.67} & 0.00 & 0.00 & 50.00 & 0.00 & 0.00 & 50.00 \\
    \midrule
    \multirow{4}[2]{*}{BLINDING} & \multirow{2}[1]{*}{VOC} & YOLOv5 & 100.00 & 0.00 & \textbf{100.00} & - & - & - & 0.00 & 0.00 & 50.00 \\
          &       & FRCNN & 100.00 & 0.00 & \textbf{100.00} & 77.78 & 0.00 & 88.89 & 0.00 & 0.00 & 50.00 \\
          & \multirow{2}[1]{*}{COCO} & YOLOv5 & 100.00 & 5.56 & \textbf{97.22} & - & - & - & 0.00 & 0.00 & 50.00 \\
          &       & FRCNN & 88.89 & 0.00 & \textbf{94.44} & 55.56 & 0.00 & 77.78 & 0.00 & 0.00 & 50.00 \\
    \midrule
    \multirow{4}[2]{*}{GMA} & \multirow{2}[1]{*}{VOC} & YOLOv5 & 100.00 & 0.00 & \textbf{100.00} & - & - & - & 0.00 & 0.00 & 50.00 \\
          &       & FRCNN & 83.33 & 0.00 & \textbf{91.67} & 88.89 & 16.67 & 86.11 & 0.00 & 0.00 & 50.00 \\
          & \multirow{2}[1]{*}{COCO} & YOLOv5 & 83.33 & 0.00 & \textbf{91.67} & - & - & - & 0.00 & 0.00 & 50.00 \\
          &       & FRCNN & 100.00 & 0.00 & \textbf{100.00} & 83.33 & 66.67 & 58.33 & 0.00 & 0.00 & 50.00 \\
    \bottomrule
    \end{tabular}}
  \label{tab:acc}
\end{table}

\cref{tab:acc} presents the detection performance of \tool against two representative baselines across three scene-level attack scenarios, two datasets, and two model architectures. \tool achieves consistently strong detection performance across all evaluated settings, averaging 96.30\% accuracy on VOC with Faster R-CNN and 96.30\% accuracy on COCO with YOLOv5, while ODSCAN collapses entirely in all scene-level settings with a constant 50\% accuracy across every configuration, and MIA shows limited and inconsistent effectiveness, averaging only 75.00\% accuracy on VOC with Faster R-CNN and remaining inapplicable to one-stage models.

\begin{table}[t]
  \centering
  \caption{Comparison of \tool and MIA on AUROC. ODSCAN is omitted as it failed across all scene-level attack scenarios. Better results per column are \textbf{bolded}. }
  \resizebox{\linewidth}{!}{
    \begin{tabular}{l|cc|cc|cc|cc|cc|cc}
    \toprule
\multirow{2}{*}{Method} 
    & \multicolumn{2}{c|}{SPONGE-VOC} & \multicolumn{2}{c|}{SPONGE-COCO} 
    & \multicolumn{2}{c|}{BLINDING-VOC} & \multicolumn{2}{c|}{BLINDING-COCO} 
    & \multicolumn{2}{c|}{GMA-VOC} & \multicolumn{2}{c}{GMA-COCO} \\
\cmidrule(lr){2-3}\cmidrule(lr){4-5}\cmidrule(lr){6-7}
\cmidrule(lr){8-9}\cmidrule(lr){10-11}\cmidrule(lr){12-13}
    & YOLOv5 & FRCNN & YOLOv5 & FRCNN & YOLOv5 & FRCNN 
    & YOLOv5 & FRCNN & YOLOv5 & FRCNN & YOLOv5 & FRCNN \\
\midrule
    \tool  & \textbf{1.00} & \textbf{0.97} & \textbf{1.00} & \textbf{0.91}
           & \textbf{1.00} & \textbf{1.00} & \textbf{0.99} & \textbf{0.94}
           & \textbf{1.00} & \textbf{0.94} & \textbf{0.92} & \textbf{1.00} \\
    MIA    & -    & 0.00   & -    & 0.00
           & -    & 0.79   & -    & 0.60
           & -    & 0.91   & -    & 0.40  \\
    \bottomrule
    \end{tabular}}
  \label{tab:auc}
\end{table}

Looking across attack scenarios and datasets, \tool attains near-perfect detection in the majority of settings, achieving 100\% accuracy on most configurations. Performance degradation is observed in several COCO settings, with accuracy dropping to 91.67\% on GMA with YOLOv5 and SPONGE with Faster R-CNN, and a non-zero FPR of 5.56\% appearing on BLINDING with YOLOv5. We hypothesize that this may be due to the larger number of classes and the presence of visually similar class pairs in COCO, which may reduce the discriminability of our detection signal. Nevertheless, \tool maintains a substantial margin over both baselines across all settings. Across architectures, \tool generalizes well to both the one-stage YOLOv5 and the two-stage Faster R-CNN. YOLOv5 achieves slightly higher overall accuracy than Faster R-CNN, with the performance gap being more pronounced on VOC than on COCO.

\cref{tab:auc} further reports AUROC scores as a threshold-independent evaluation of discrimination performance. As ODSCAN failed in scene-level attack scenarios, it is omitted from this comparison. \tool achieves near-perfect AUROC across the majority of settings, reaching 1.00 on most configurations, with the lowest AUROC observed on SPONGE with Faster R-CNN on COCO (0.91). MIA's AUROC scores vary widely across settings, ranging from 0.91 on GMA with Faster R-CNN on VOC to 0.00 on SPONGE and 0.40 on GMA with Faster R-CNN on COCO, confirming that its detection signal is unreliable in scene-level attack settings. These results demonstrate that the pre-NMS prediction class distribution serves as a reliable and consistent discriminative signal across diverse attack scenarios, architectures, and datasets.

\noindent\textbf{Generalization to Additional Attacks and Architectures.} 
We also test \tool on VOC with BadDet OGA, RMA, and ODA~\cite{chan2022baddet}, plus YOLOv8 and DETR, adding 360 models over 18 new settings. \tool reaches a mean AUROC of 0.87, outperforming ODSCAN (0.58) and MIA (0.73), and achieves the best AUROC in every new setting. Detailed per-setting AUROCs are reported in the appendix.

\noindent\textbf{Analysis of Baseline Failures.} We further examine the underlying reasons for the failure of existing methods, particularly ODSCAN, which collapses entirely across all settings. ODSCAN is designed around the assumption that a backdoor manifests as a localized trigger inducing a specific victim-to-target class transition, an assumption that is directly violated by scene-level attacks, where no specific victim class exists (\ie, BLINDING and GMA affect all classes indiscriminately) and no specific target class exists (\ie, SPONGE generates arbitrary false positives). Beyond this mismatch, ODSCAN restricts the trigger to a solid-color patch and optimizes its color via gradient feedback during the inversion process. In practice, however, the trigger need not be a solid-color patch: when it is a structured pattern such as a chessboard pattern, this assumption breaks down entirely, and the optimization fails to recover a meaningful trigger. Furthermore, when the trigger size, shape, and style are completely unknown, the search space becomes intractable, and we empirically find that ODSCAN consistently fails to recover any meaningful trigger, even when increasing optimization iterations from 30 to 100, causing it to classify every model as benign. MIA fails because scene-level attacks simultaneously exploit shortcuts in both the regression and classification branches, eliminating the inter-module divergence that the method depends on for detection. \tool sidesteps both limitations by operating on a distributional signal derived from clean inputs, requiring neither assumptions about trigger form nor architectural constraints, and capturing any systematic distortion of the model's class-level behavior regardless of the attack mechanism or model architecture.

\subsubsection{Effect of Validation Set Construction.}
\label{sec:ablation_valset}

\begin{table}[t]
  \centering
  \caption{Detection accuracy of \tool under different validation set construction strategies. Best results per column are \textbf{bolded}. \colorbox{blue!12}{Shaded rows} indicate the construction strategy used in this setting. }
    \resizebox{\linewidth}{!}{
    \begin{tabular}{cc|ccc|c|ccc|c}
    \toprule
    \multirow{2}[4]{*}{Dataset} & \multirow{2}[4]{*}{Val Set} & \multicolumn{4}{c|}{YOLOv5} & \multicolumn{4}{c}{Faster R-CNN} \\
\cmidrule(lr){3-6}\cmidrule(lr){7-10}
          &       & SPONGE & BLINDING & GMA   & Average   & SPONGE & BLINDING & GMA   & Average \\
    \midrule
    \multirow{3}[1]{*}{VOC} & Equal Number & \textbf{100.00} & \textbf{100.00} & \textbf{100.00} & \textbf{100.00} & 55.56  & 69.44  & 77.78  & 67.59  \\
          & Equal Size   & \textbf{100.00} & 94.44  & 72.22  & 88.89  & \cellcolor{blue!10}\textbf{97.22}  & \cellcolor{blue!10}\textbf{100.00} & \cellcolor{blue!10}\textbf{91.67}  & \cellcolor{blue!10}\textbf{96.30}  \\
          & Random       & \cellcolor{blue!10}\textbf{100.00} & \cellcolor{blue!10}\textbf{100.00} & \cellcolor{blue!10}\textbf{100.00} & \cellcolor{blue!10}\textbf{100.00} & 55.56  & 69.44  & 77.78  & 67.59  \\
    \midrule
    \multirow{3}[1]{*}{COCO} & Equal Number & \textbf{100.00} & 94.44  & 86.11  & 93.52  & 55.56  & 88.89  & 88.89  & 77.78  \\
          & Equal Size   & \textbf{100.00} & 88.89  & \textbf{97.22}  & 95.37  & \cellcolor{blue!10}\textbf{91.67}  & \cellcolor{blue!10}\textbf{94.44}  & \cellcolor{blue!10}\textbf{100.00} & \cellcolor{blue!10}\textbf{95.37}  \\
          & Random       & \cellcolor{blue!10}\textbf{100.00} & \cellcolor{blue!10}\textbf{97.22}  & \cellcolor{blue!10}91.67  & \cellcolor{blue!10}\textbf{96.30}  & 55.56  & 72.22  & 77.78  & 68.52  \\
    \bottomrule
    \end{tabular}}%
  \label{tab:diff_eval}%
\end{table}

We evaluate how different validation set construction strategies affect the performance of \tool. Three strategies are considered. \textit{Equal Number} samples images directly from the original dataset while ensuring that the total number of object instances per category is fixed at 10. \textit{Equal Size} crops one selected object per image using its ground-truth bounding box, resizes crops to the median crop size across all samples, and keeps the same per-category instance count; this is the strategy adopted by \tool for Faster R-CNN. \textit{Random} samples images from the original dataset without any control over category balance or image content, with the total image count matched to the Equal Number setting; this is the strategy adopted by \tool for YOLOv5.

As shown in~\cref{tab:diff_eval}, the two architectures respond differently to these strategies, in a manner consistent with how each model's classification head is trained. For YOLOv5, \textit{Equal Number} and \textit{Random} consistently achieve strong performance, both reaching 100.00\% on all attack scenarios on VOC, while \textit{Equal Size} degrades on BLINDING and GMA. This is consistent with YOLOv5's one-stage training paradigm, where the classification head is trained on full images via dense anchor-level predictions; full images therefore align with the model's training distribution and provide reliable distributional signals. On COCO, \textit{Random} achieves slightly higher overall accuracy than \textit{Equal Number} (96.30\% vs.\ 93.52\%), and we therefore adopt \textit{Random} as the default validation set construction strategy for YOLOv5.

For Faster R-CNN, the pattern reverses: \textit{Equal Number} and \textit{Random} perform substantially worse, while \textit{Equal Size} consistently yields the highest detection accuracy, with a particularly notable improvement on SPONGE (97.22\% vs.\ 55.56\% on VOC). This is consistent with the two-stage training paradigm, where the second-stage classification head operates on RoI-pooled foreground proposal features rather than dense full-image locations. Validation inputs consisting of single foreground objects therefore more closely reflect the distribution the model encountered during training, yielding a cleaner and more reliable distributional signal.

Furthermore, for YOLOv5, \textit{Equal Number} and \textit{Random} achieve comparable performance despite differing in category balance control, suggesting that explicitly enforcing per-category instance counts provides no additional benefit for one-stage models. This implies that \tool's distributional signal is robust to validation set composition: as long as inputs match the model's training distribution, the attack-induced shift in class distribution manifests reliably, whether the validation set is balanced or randomly sampled.

\subsubsection{Effect of Confidence Threshold.}
\label{sec:ablation_para}

\begin{table}[t]
  \centering
  \caption{Detection Accuracy of \tool using different confidence thresholds $\delta$. 
           Best results per column are \textbf{bolded}. 
           \colorbox{blue!12}{Shaded rows} indicate the recommended threshold.}
  \resizebox{\linewidth}{!}{
  \begin{tabular}{cc|ccc|c|ccc|c}
    \toprule
    \multirow{2}[4]{*}{Dataset} & \multirow{2}[4]{*}{$\delta$} 
      & \multicolumn{4}{c|}{YOLOv5} & \multicolumn{4}{c}{Faster R-CNN} \\
    \cmidrule(lr){3-6}\cmidrule(lr){7-10}
      & & SPONGE & BLINDING & GMA & \textbf{All} 
        & SPONGE & BLINDING & GMA & \textbf{All} \\
    \midrule
    \multirow{5}[2]{*}{VOC}
      & 0.0000 & \textbf{100.00} & \textbf{100.00} & \textbf{100.00} & \textbf{100.00}
               & 50.00   & 50.00   & 50.00   & 50.00   \\
    \rowcolor{blue!10}
      \cellcolor{white} & 0.0005 & \textbf{100.00} & \textbf{100.00} & \textbf{100.00} & \textbf{100.00}
               & \textbf{97.22}  & \textbf{100.00} & \textbf{91.67}  & \textbf{96.30} \\
      & 0.0010 & \textbf{100.00} & 97.22  & 91.67  & 96.30
               & \textbf{97.22}  & \textbf{100.00} & \textbf{91.67}  & \textbf{96.30} \\
      & 0.0100 & \textbf{100.00} & 97.22  & 77.78  & 91.67
               & 50.00  & 86.11  & 91.67  & 75.93  \\
      & 0.0500 & 94.44  & 97.22  & 80.56  & 90.74
               & 52.78  & 72.22  & 88.89  & 71.30  \\
    \midrule
    \multirow{5}[2]{*}{COCO}
      & 0.0000 & \textbf{100.00} & 97.22  & \textbf{91.67}  & 96.30
               & 50.00   & 50.00   & 50.00   & 50.00   \\
    \rowcolor{blue!10}
     \cellcolor{white} & 0.0005 & \textbf{100.00} & 97.22  & \textbf{91.67}  & 96.30
               & \textbf{91.67}  & \textbf{94.44}  & \textbf{100.00} & \textbf{95.37} \\
      & 0.0010 & \textbf{100.00} & 97.22  & \textbf{91.67}  & 96.30
               & 55.56  & 86.11  & 61.11  & 67.59  \\
      & 0.0100 & \textbf{100.00} & \textbf{100.00} & \textbf{91.67} & \textbf{97.22}
               & 66.67  & 80.56  & 80.56  & 75.93  \\
      & 0.0500 & 97.22  & \textbf{100.00} & 88.89  & 95.37
               & 69.44  & 80.56  & 86.11  & 78.70  \\
    \bottomrule
  \end{tabular}}
  \label{tab:ablation_threshold}
\end{table}

We investigate the effect of the confidence threshold $\delta$, which filters out low-confidence intermediate predictions before constructing the empirical class distribution. We evaluate five values of $\delta \in \{0.0000, 0.0005, 0.0010, 0.0100, 0.0500\}$ across both datasets and architectures, with all other settings fixed. The value used in our main experiments is $\delta = 0.0005$.

\cref{tab:ablation_threshold} reveals two distinct failure modes at the extremes of $\delta$. When $\delta = 0.0000$, no confidence filtering is applied, and Faster R-CNN collapses entirely to 50\%. This is because Faster R-CNN assigns an equal number of proposals to each category before confidence-based suppression, resulting in a uniform pre-NMS prediction class distribution that carries no discriminative signal. As $\delta$ increases beyond 0.0010, performance degrades progressively on both architectures, indicating that overly aggressive filtering discards genuine object predictions along with noise and weakens the distributional signal. An exception is observed for YOLOv5 on COCO, where $\delta = 0.0100$ yields slightly higher accuracy (97.22\%) than $\delta = 0.0005$ (96.30\%), though the difference is marginal. Considering overall performance across all architectures and datasets, $\delta = 0.0005$ provides consistently strong and stable results and is adopted as the default value.

\begin{figure}[tb]
  \centering
  \includegraphics[width=\linewidth]{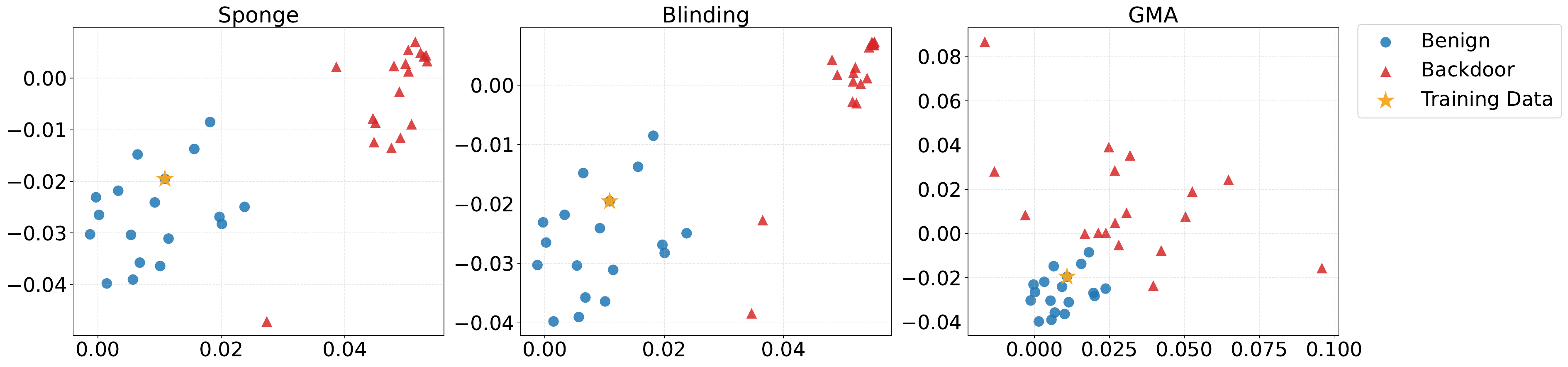}
  \caption{PCA projection of pre-NMS prediction class distributions for benign (blue circles) and backdoored (red triangles) models across three attack scenarios on Faster R-CNN, with the training data distribution marked as a yellow star.
  }
  \label{fig:distribution_frcnn}
\end{figure}

\subsubsection{Visualization of Distribution Shift.}

To provide an intuitive understanding of the distributional signal that \tool exploits, we project the empirical class distributions extracted from benign models, backdoored models, and the training data onto a two-dimensional space via PCA~\cite{mackiewicz1993principal}, with results shown in~\cref{fig:distribution_frcnn}.

Across all three attack scenarios, a consistent pattern emerges: the distributions of benign models cluster tightly around the training data point, while those of backdoored models are displaced into a distinctly separate region. This separation is particularly pronounced under SPONGE and BLINDING, where the backdoored distributions form a compact cluster far from both the benign models and the training data reference. Under GMA, the backdoored distributions are more spread out, reflecting the greater variability in how this attack distorts class-level predictions across different model initializations, yet they remain clearly separable from the benign cluster. These visualizations directly support the core premise of \tool: backdoor training shifts the intermediate class prediction distribution away from the training data statistics, and this shift is detectable from clean inputs alone without any knowledge of the trigger.

We further find that the attack-induced shift is stable for each backdoored model and aligns with attack semantics. Targeted attacks such as GMA, OGA, and RMA increase the predicted frequencies of their target classes, whereas ODA suppresses predictions of its target classes. In contrast, SPONGE and BLINDING produce broader multi-class shifts. This supports that \tool captures a systematic model-level signature rather than random per-image fluctuations. We provide a more detailed characterization in the appendix.

\section{Conclusion}

In this paper, we presented \tool, a backdoor detection framework for object detection models. Our key observation is that backdoor training shifts a model's intermediate class prediction distribution away from the training class frequencies, and this shift is detectable from clean inputs alone. By measuring the Jensen-Shannon divergence between the empirical class distribution extracted from a small constructed validation set and the expected training class frequencies, \tool provides a simple yet effective detection signal that requires no model weight access, no trigger knowledge, and no additional training. Extensive experiments demonstrate that \tool substantially outperforms existing methods. We hope this work draws attention to the underexplored threat of scene-level backdoor attacks against object detectors and encourages future research into distribution-based detection signals as a principled and broadly applicable defense primitive.

\section*{Acknowledgements}
This work was supported by the New Generation Artificial Intelligence-National Science and Technology Major Project (2025ZD0123305), the National Key Research and Development Program of China (2023YFB3107401), the National Science and Technology Major Project of the Ministry of Science and Technology of China (2025ZD0805904), the National Natural Science Foundation of China (T2341003, 62521002, U2441240, U24B20185, 62376210, 62132011, 62406240), Xinjiang Tianshan Innovative Research Team (2025D14009), Shaanxi Provincial Key R\&D Program (Key Projects 2025ZG-JBGS-001), the National Research Foundation, Singapore and CyberSG R\&D Programme Office under its Translation and Innovation Grant (CRPO-GC4-SMU-002). Any opinions, findings and conclusions or recommendations expressed in this material are those of the author(s) and do not reflect the views of the National Research Foundation and CyberSG R\&D Programme Office.

\begingroup
\small
\bibliographystyle{unsrt}
\bibliography{main}
\endgroup
\end{document}